\documentclass{article}

\usepackage{microtype}
\usepackage{graphicx}
\usepackage{subfigure}
\usepackage{booktabs} 

\usepackage{hyperref}

\usepackage[accepted]{icml2023}

\usepackage{amsmath}
\usepackage{amssymb}
\usepackage{mathtools}
\usepackage{amsthm}

\usepackage[capitalize,noabbrev]{cleveref}

\theoremstyle{plain}

\theoremstyle{definition}

\theoremstyle{remark}

\usepackage[textsize=tiny]{todonotes}

\icmltitlerunning{The Forward-Forward Algorithm as a feature extractor for skin lesion classification: A preliminary study.}

\begin{document}

\twocolumn[
\icmltitle{The Forward-Forward Algorithm as a feature extractor for skin lesion classification: A preliminary study}

\icmlsetsymbol{equal}{*}

\begin{icmlauthorlist}
\icmlauthor{Abel Reyes-Angulo}{mtu}
\icmlauthor{Sidike Paheding}{mtu}

\end{icmlauthorlist}

\icmlaffiliation{mtu}{Department of Applied Computing, Michigan Technological University, Houghton, USA}

\icmlcorrespondingauthor{Abel Reyes-Angulo}{areyesan@mtu.edu}
\icmlcorrespondingauthor{Sidike Paheding}{spahedin@mtu.edu}

\icmlkeywords{Machine Learning, medical imaging, learning representation}

\vskip 0.3in
]

\printAffiliationsAndNotice

\begin{abstract}

Skin cancer, a deadly form of cancer, exhibits a 23\% survival rate in the USA with late diagnosis. Early detection can significantly increase the survival rate, and facilitate timely treatment. Accurate biomedical image classification is vital in medical analysis, aiding clinicians in disease diagnosis and treatment. Deep learning (DL) techniques, such as convolutional neural networks and transformers, have revolutionized clinical decision-making automation. However, computational cost and hardware constraints limit the implementation of state-of-the-art DL architectures.
In this work, we explore a new type of neural network that does not need backpropagation (BP), namely the Forward-Forward Algorithm (FFA), for skin lesion classification. While FFA is claimed to use very low-power analog hardware,  BP still tends to be superior in terms of  classification accuracy. In addition, our experimental results suggest that the combination of FFA and BP can be a better alternative to achieve a more accurate prediction.

\end{abstract}

\section{Introduction}\label{sec1}
Skin cancer is one of the most common types of cancer \cite{linares2015skin, siegel2023cancer}, and early detection and diagnosis can greatly improve the prognosis of patients \cite{miller2019cancer}. Dermatologists use visual inspection to diagnose skin cancer, which can be challenging because there are numerous skin lesions and the visual similarity between different types of them. Therefore, Computer-aided diagnosis (CAD) systems are required to assist dermatologists in the early detection and diagnosis of skin cancer, aiming to ensure timely treatment \cite{bhinder2021artificial}.
In the past few years, deep learning (DL) has revolutionized the automation of clinical decisions through computational system assistance. Techniques such as backpropagation \cite{rumelhart1985learning}, convolutional neural networks (CNNs) \cite{lecun1998gradient}, and recently proposed transformers \cite{vaswani2017attention} have boosted confidence in the use of AI in a real clinical setup. However, the computational cost and hardware constraints related to the implementation of state-of-the-art DL architectures limit the feasibility of actual deployment implementation \cite{imteaj2022federated}.
The Forward-Forward Algorithm (FFA) \cite{hinton2022forward} has been presented as an alternative to optimizing the training process of neural networks, considering the ``mortal computational" cost. Although results reported by the use of FFA have not outperformed traditional mechanisms such as backpropagation, the trade-off could be less requirement in hardware implementation.
In this work, we explore the performance of FFA for skin lesion classification and also compare it to a backpropagation-based trained model. Finally, we propose to combine the two techniques during the training process, where the FFA is used as a feature extractor and backpropagation to improve model prediction. 

\section{Related works}\label{sec2}
Skin lesion classification has been a significant research area in recent years, and numerous deep-learning techniques have been designed for this task \cite{zhang2019attention, mahbod2019skin}. Most existing methods use traditional CNNs, which learn the feature representation through backpropagation \cite{harangi2018skin, albahar2019skin}. However, these methods can be limited by the quality of the feature representation.
In order to overcome this limitation, some researchers have suggested methods that use adversarial training to improve the feature representation. For example, GAN-based methods have been implemented for skin lesion classification, which use a generator network to generate realistic skin lesion images and a discriminator network to classify the generated images as real or fake \cite{rashid2019skin}.

In contrast to these methods, we herein propose to combine the FFA with traditional backpropagation to enhance feature representation and improve classification accuracy in skin lesion classification. Our approach does not require adversarial training or pre-training on a large dataset, making it computationally efficient and suitable for large-scale datasets. Our approach means to provide a baseline on the performance of DL architecture using the FFA strategy in contrast with backpropagation, in addition to exploring the combination of the two techniques.

\section{Methodology}\label{sec3}
Our proposed framework consists of two stages: feature representation with FFA and final prediction with backpropagation.

\subsection{Backpropagation}
Introduced by Rumelhart et al. \cite{rumelhart1986learning, rumelhart1985learning}, 
the backpropagation algorithm, utilized for training artificial neural networks, entails calculating the gradient of the error function concerning the network weights. This gradient information is then used to update the weights by using gradient descent, aiming to minimize the global loss function of the task. The authors demonstrated that this algorithm could facilitate the learning of distributed representation of input data, enabling effective generalization to new data. 

The backpropagation algorithm consists of two main stages: the forward pass and the backward pass. During the forward pass, input data is propagated through the network, generating corresponding predictions. Subsequently, during the backward pass, the gradients of the loss function  are calculated with respect to the network's parameters by propagating the error backward through the neural network. These gradients are then utilized to update the parameters using an optimization algorithm, such as stochastic gradient descent (SGD) or Adam \cite{amari1993backpropagation, bock2018improvement}.

\subsection{The Forward-Forward Algorithm}

\begin{figure}[h]%
\centering
\includegraphics[width=0.5\textwidth]{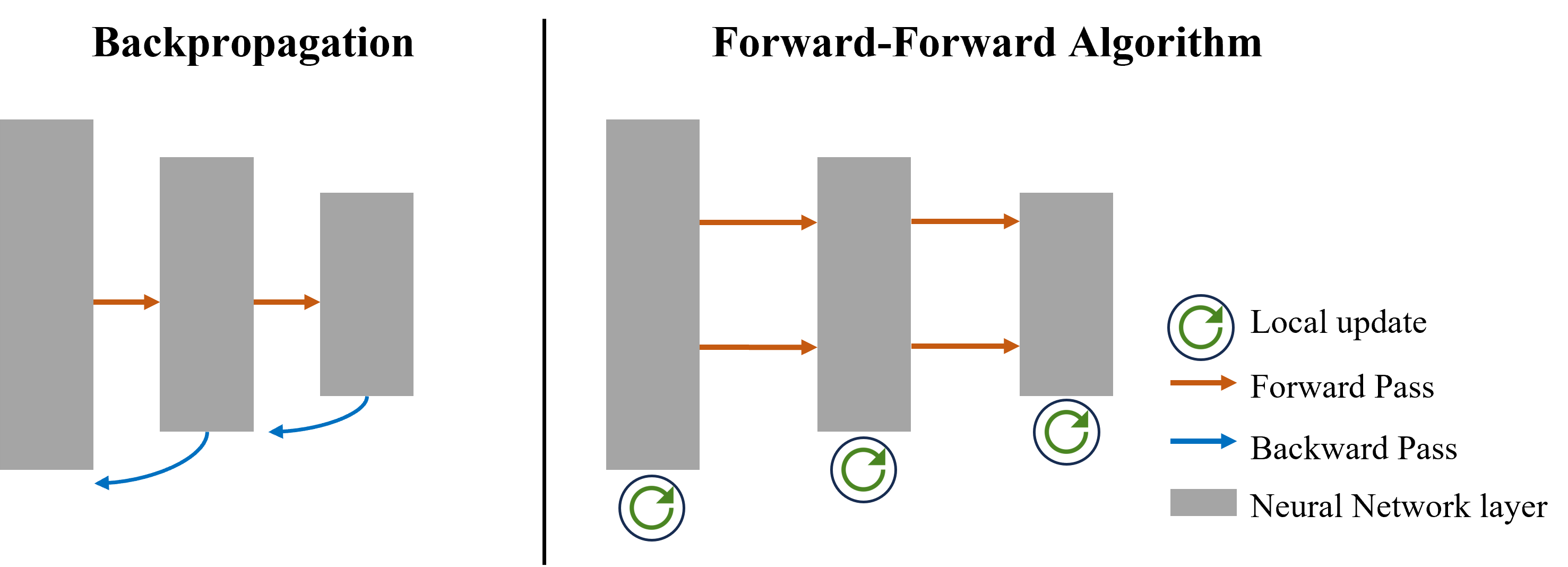}
\caption{A brief illustration of the backpropagation and the forward-forward algorithm.}\label{fig1}
\end{figure}

Introduced by Geoffrey Hinton \cite{hinton2022forward}, the forward-forward algorithm (FFA) is an innovative learning procedure for neural networks. It draws inspiration from the Boltzmann machines \cite{hinton1986learning} and Noise Contrastive Estimation \cite{gutmann2010noise}, Its primary objective is  to replace the traditional forward and backward passes of backpropagation with two parallel forward passes. The first forward pass uses real data, and the second forward pass uses negative data, which the network generates itself. Each layer possesses its individual objective function, which essentially aims to maximize the goodness for positive data while minimizing the goodness for negative data. Figure \ref{fig1} illustrates the main difference between the traditional backpropagation algorithm and the newer FFA. Mathematically, the goodness is computed as follows
\begin{equation}
    prob(\text{positive-data})=\sigma(\sum_j y_j^2-\theta))
\end{equation}
where $y_i$ represents activity of the $j^{th}$ hidden unit, $\theta$ a given threshold, and $\sigma$ a logistic distribution function. 
The FFA has been shown to work well on a few small problems \cite{ororbia2023learning}, but it has not yet been tested on large-scale problems.  In addition, the FFA has been claimed to be superior in hardware efficiency, through low power consumption when compared with the backpropagation and the gradient computation \cite{hinton2022forward, kendall2020training}.

\subsection{Feature representation in FFA}

In the FFA, pairs of data points are created by corrupting real data points with noise. The network is then trained to distinguish between the real data points and the corrupted data points. This training process forces the network to learn representations that are robust to noise.
The proposed method utilizes the FFA algorithm as the initial stage of learning, and then the final classification is trained by using regular backpropagation.

\section{Implementation details}\label{sec4}
\begin{figure}[h]%
\centering
\includegraphics[width=0.46\textwidth]{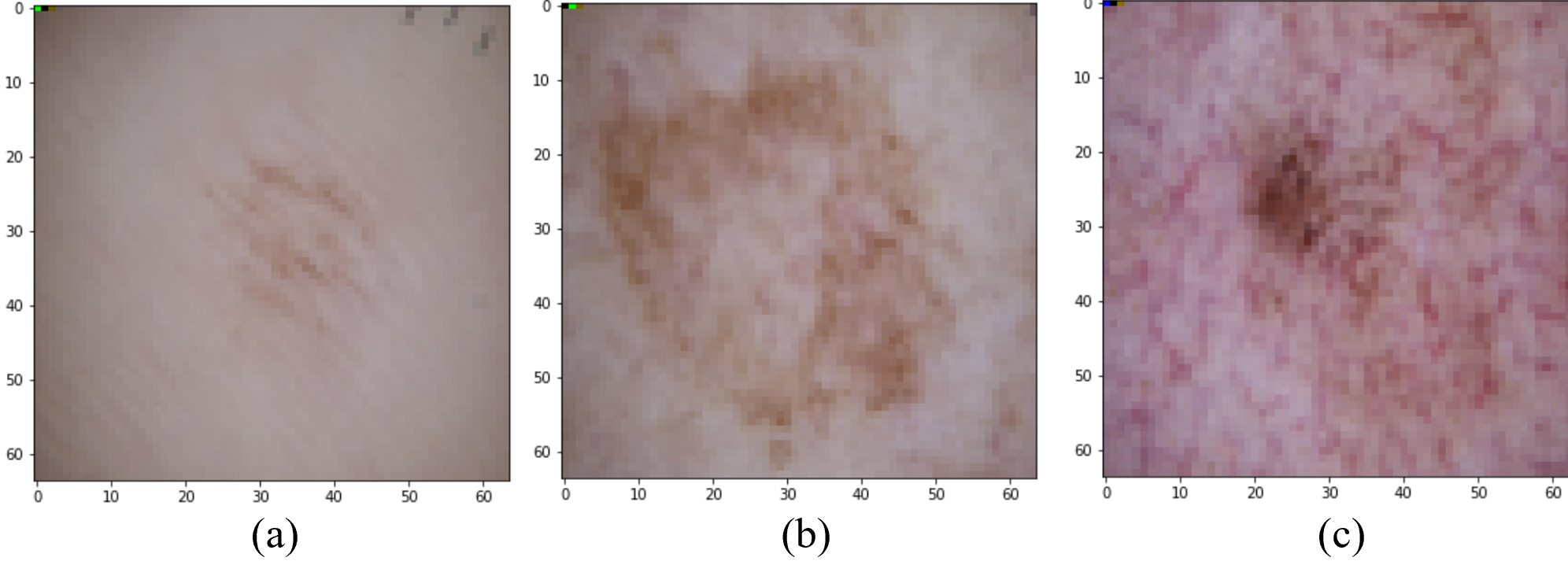}
\caption{Illustration of the generated positive and negative data when FFA is used over the HAM 10000 dataset. The images contain embedded label information in the first 7 pixels: (a) sample with label 1, (b) sample with label 4, and (c) sample with the label 2.}\label{fig2}
\end{figure}

The proposed methodology was tested with two skin lesion classification benchmark datasets: ISIC 2016 and HAM 10000 datasets, details about the dataset are provided in section \ref{sec5}. The ISIC 2016 provides training and testing datasets with the correspondent classification ground truth (i.e. benign or malign skin lesion). However, the HAM 10000 dataset provides only the training dataset with the corresponding type of skin lesion ground truth. For the last case, we randomly split the dataset in training and testing with a ratio of 8:2, respectively. In order to alleviate the computational constraints all the images, from the aforementioned datasets, were downsampled to $64\times64$ pixels size, maintaining the original RGB channels. The inputs were flattened and normalized prior to the training process. The architecture for all the experiments was composed of 3 fully connected hidden layers (784, 500, 500 units) with ReLU activation and batch normalization. A final softmax activation function provides the probability distribution among the classes corresponding to the classification task on each dataset. 
In the case of the FFA, the positive and negative data are generated by the network, and the one-hot-encoder representation of the label is overlying in the first $n$ pixels of the input image, where $n$ represent the number of classes available. The positive data embed the original ground truth in the corresponding input image; meanwhile, the negative data overlay a wrong label in the same input image. Figure \ref{fig2} illustrates the overlay technique utilized in this work to generate the positive and the negative data. The architecture is implemented within a Tensorflow/Keras 2.9 version environment. Each experiment is run using a single NVIDIA GeForce RTX 3070 graphic card with 8GB of dedicated GPU. All experiments are run for a total of 250 epochs with checkpoint callback, and batch size of 64 samples per iteration. The mean squared error (MSE) loss was used along with the Adam optimizer with an initial learning rate of $1 \times 10^{-3}$.

\section{Experiments}\label{sec5}
In this section, we provide more details about the experiments performed to evaluate the proposed method, including the description of the datasets, performance metrics used as evaluation criteria, and a discussion of the experimental results.

\subsection{Datasets}
In order to evaluate the robustness of the proposed methodology, we utilize two widely used and publicly available benchmark datasets related to skin lesion classification: ISIC 2016 and HAM 10000 datasets.

\begin{figure}[h]%
\centering
\includegraphics[width=0.45\textwidth]{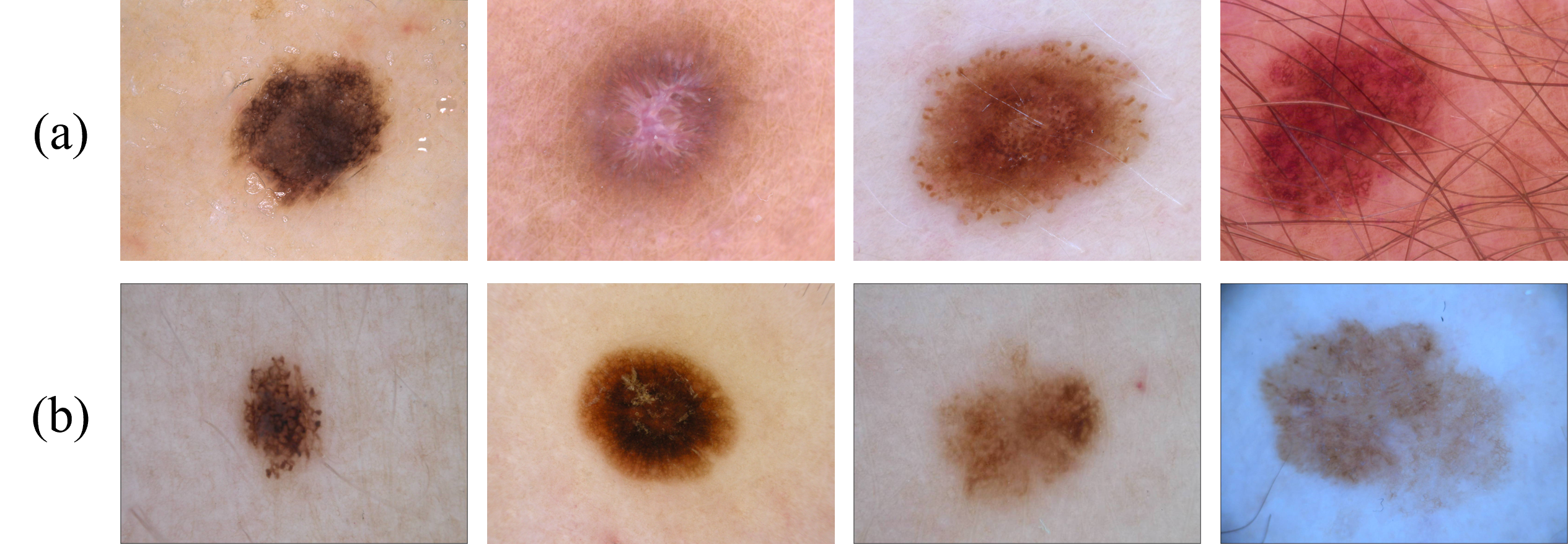}
\caption{Samples from the two benchmark datasets utilized in the work: (a) The ISIC 2016 and (b) the HAM 10000 dataset.}\label{fig3}
\end{figure}

\subsubsection{ISIC 2016}
Presented in the International Skin Imaging Collaboration (ISIC) 2016 challenge, the ISIC 2016 dataset \cite{gutman2016skin} is a comprehensive collection of dermoscopic images that is widely used in the field of computer-aided diagnosis for skin lesions. It consists of 900 training images and 379 test images of dermoscopic skin lesions. The dataset includes binary masks for both lesion segmentation and classification purposes, enabling researchers to evaluate algorithms for these tasks.
The dataset encompasses three main types of skin lesions: melanoma (Malign), seborrheic keratosis, and nevus (Benign). These lesion types cover a range of conditions commonly encountered in dermatology practice. The images in the dataset were obtained from patients across multiple countries, resulting in a diverse and challenging dataset.

\subsubsection{HAM 10000}
\begin{figure}[h]%
\centering
\includegraphics[width=0.48\textwidth]{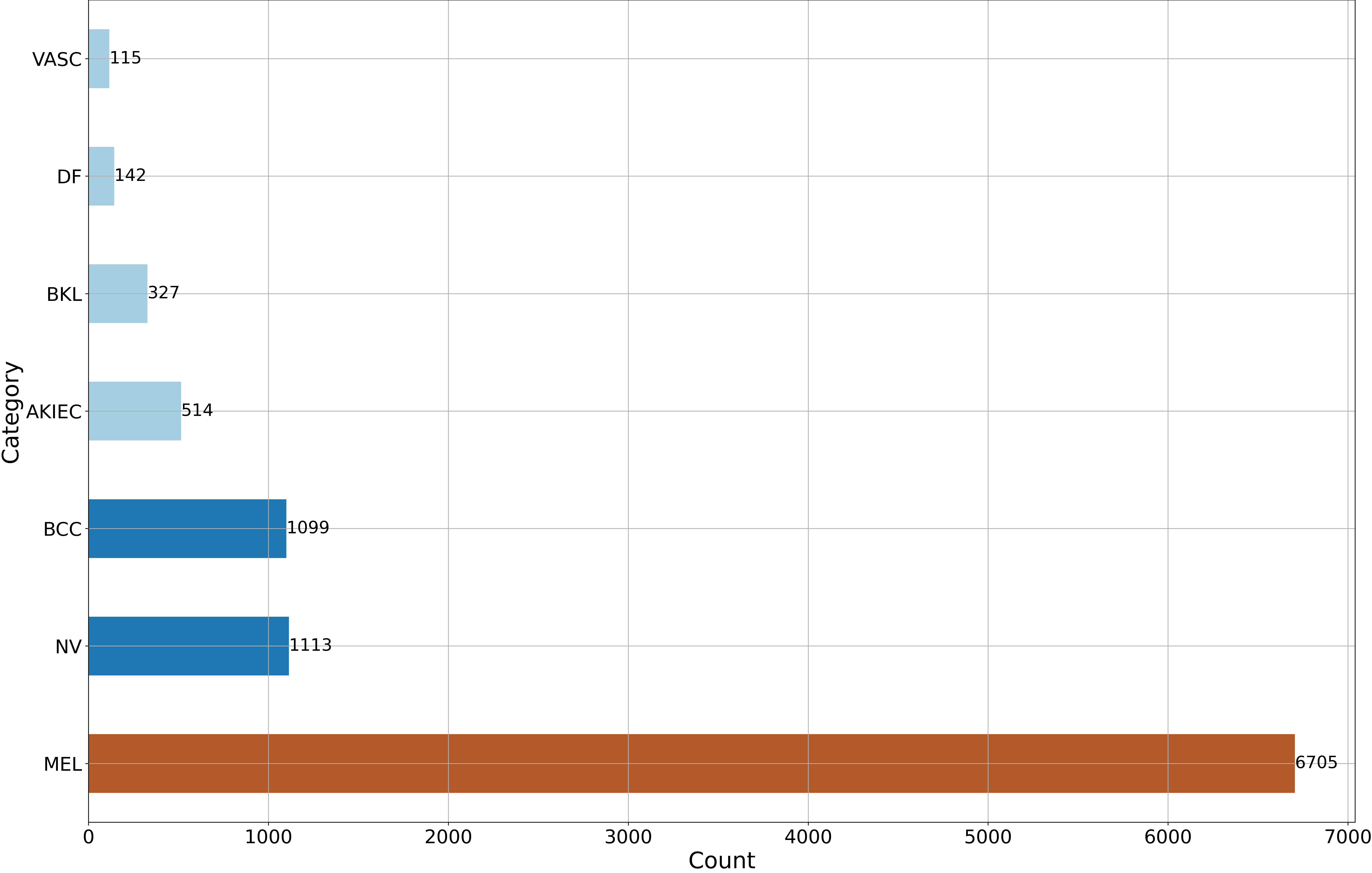}
\caption{Distribution of the samples per class in the HAM 10000 dataset.}\label{fig5}
\end{figure}

The Human Against Machine with 10000 training images (HAM10000) dataset \cite{tschandl2018ham10000}, widely used in dermatology and computer vision, comprises 10,015 dermoscopic images of skin lesions obtained from diverse clinical sources. It was specifically created for the ISIC 2018 challenge initiated by the International Skin Imaging Collaboration, with the aim of advancing automated analysis of skin lesions.
Within the HAM10000 dataset, images represent seven distinct types of skin lesions: melanoma(MEL), melanocytic nevus(NV), basal cell carcinoma(BCC), actinic keratosis(AKIEC), benign keratosis(BLK), dermatofibroma(DF), and vascular lesions(VASC). Figure \ref{fig5} shows the numerical by-class distribution of the samples in the HAM 100000 dataset. 
Figure \ref{fig3} shows samples of the two mentioned datasets.

\subsection{Performance metrics}

\subsubsection{the error rate \%:} The error rate is a commonly used performance metric that measures the accuracy of the model's predictions. It represents the proportion of incorrectly classified instances out of the total number of instances. The error rate, denoted as $\epsilon$, is a straightforward and intuitive metric that provides a simple measure of the model's performance in image classification tasks and can be mathematically expressed as
\begin{equation}
    \epsilon = \frac{FP + FN}{TP + TN + FP + FN} \times 100\%
\end{equation}

\subsubsection{The ROC-AUC score:} The ROC-AUC measures the classifier's ability to distinguish between positive and negative instances across various classification thresholds. The ROC curve (receiver operating characteristic curve) demonstrates the true positive rate (sensitivity) against the false positive rate (1-specificity), and the ROC-AUC represents the area under this ROC curve. A higher ROC-AUC indicates better classification performance. The ROC-AUC can be mathematically expressed as
\begin{equation}
    \text{ROC AUC score} = \int_{-\infty}^{\infty} \text{TPR}(f) \cdot \text{FPR}(f) 
    \text{d}f
\end{equation}

where $\text{{TPR}}(f) = \frac{{\text{{TP}}(f)}}{{\text{{P}}}}$ and $\text{{FPR}}(f) = \frac{{\text{{FP}}(f)}}{{\text{{N}}}}$ are knowing as the true positive rate and the false positive rate, respectively

\subsubsection{Experimental results and discussion}
Experimental results are reported in Table \ref{tab1} in terms of the error rate \%, and in Table\ref{tab2} in terms of ROC-AUC score. In both cases, it shows superior performance of the model when it is trained with the backpropagation method in contrast when it is trained with FFA. However, reported results suggest a better performance can be achieved when both techniques are combined, and the results are consistent over the two datasets.
It is important to remark on the following points: (1) The architecture configuration was not intended to achieve state-of-the-art performance but provides a comparative baseline when a DL architecture is trained using either FFA, backpropagation, or both. (2) The use of only fully connected layers limits the size of the architecture due to computational constraints. Therefore, this is expected to be addressed through the implementation of convolutional layers.

\begin{table}[]
\caption{Error rate (\%) on skin lesion datasets.}
\label{tab1}
\centering
\scriptsize 
\begin{tabular}{l|cc|cc|}
\cline{2-5}
                             & \multicolumn{2}{c|}{ISIC 2016} & \multicolumn{2}{c|}{HAM 10000} \\ \cline{2-5} 
 &
  \multicolumn{1}{r|}{\begin{tabular}[c]{@{}r@{}}Testing error\\ rate \%\end{tabular}} &
  \multicolumn{1}{r|}{\begin{tabular}[c]{@{}r@{}}Training error\\ rate \%\end{tabular}} &
  \multicolumn{1}{r|}{\begin{tabular}[c]{@{}r@{}}Testing error\\ rate \%\end{tabular}} &
  \multicolumn{1}{r|}{\begin{tabular}[c]{@{}r@{}}Training error\\ rate \%\end{tabular}} \\ \hline
\multicolumn{1}{|l|}{FFA}    & \multicolumn{1}{c|}{29.52} & 22.48 & \multicolumn{1}{c|}{37.21} & 20.42 \\ \hline
\multicolumn{1}{|l|}{BP}     & \multicolumn{1}{c|}{24.81} & 17.33 & \multicolumn{1}{c|}{31.66} & 13.97 \\ \hline
\multicolumn{1}{|l|}{FFA+BP} & \multicolumn{1}{c|}{\textbf{23.31}} & \textbf{16.85} & \multicolumn{1}{c|}{\textbf{30.44}} & \textbf{13.43}\\ \hline
\end{tabular}
\end{table}

\begin{table}[]
\caption{ROC-AUC on skin lesion classification benchmarks.}
\label{tab2}
\centering
\scriptsize 
\begin{tabular}{l|cc|cc|}
\cline{2-5}
                             & \multicolumn{2}{c|}{ISIC 2016} & \multicolumn{2}{c|}{HAM 10000} \\ \cline{2-5} 
 &
  \multicolumn{1}{r|}{\begin{tabular}[c]{@{}r@{}} Testing\\ ROC-AUC \end{tabular}} &
  \multicolumn{1}{r|}{\begin{tabular}[c]{@{}r@{}} Training\\ ROC-AUC \end{tabular}} &
  \multicolumn{1}{r|}{\begin{tabular}[c]{@{}r@{}} Testing\\ ROC-AUC \end{tabular}} &
  \multicolumn{1}{r|}{\begin{tabular}[c]{@{}r@{}} Training\\ ROC-AUC \end{tabular}} \\ \hline
\multicolumn{1}{|l|}{FFA}    & \multicolumn{1}{c|}{0.5632} & 0.6684 & \multicolumn{1}{c|}{0.6912} & 0.7824 \\ \hline
\multicolumn{1}{|l|}{BP}     & \multicolumn{1}{c|}{0.6216} & 0.8240 & \multicolumn{1}{c|}{0.8360} & 0.9324 \\ \hline
\multicolumn{1}{|l|}{FFA+BP} & \multicolumn{1}{c|}{\textbf{0.6495}} & \textbf{0.8472} & \multicolumn{1}{c|}{\textbf{0.8483}} & \textbf{0.9414} \\ \hline
\end{tabular}
\end{table}

\section{Conclusion}\label{sec7}
In this work, we investigated the performance of FFA for the skin lesion classification task. We compared the prediction accuracy of the FFA, backpropagation, and the combination of both. Experimental results showed that the backpropagation algorithm yields better classification accuracy than FFA. However, our results also suggest promising insights into the use of the FFA in the neural network as a feature extractor. A contrastive learning stage with FFA complemented with classification learning using traditional backpropagation could enhance the performance for skin lesion classification when comparing the similar architecture setup using either FFA or the backpropagation method only.  One limitation of our study is that we did not include the results of the energy efficiency of FFA in terms of hardware implementation, which will be our future work.

\nocite{langley00}

\bibliography{references}
\bibliographystyle{icml2023}

\end{document}